\documentclass[letterpaper]{article}
\usepackage{aaai}
\usepackage{times}
\usepackage{helvet}
\usepackage{courier}
\usepackage[linesnumbered,ruled]{algorithm2e}
\let\chapter\undefined
\usepackage{varioref}
\usepackage{graphicx}
\usepackage{subfig}
\usepackage{amsmath}
\usepackage{amsthm}
\usepackage{amssymb}
\usepackage{paralist}
\usepackage{xcolor}
\usepackage{soul}
\usepackage{comment}
\usepackage{tabularx}
\usepackage{CJK}
\theoremstyle{definition}
\newtheorem{definition}{Definition}
\theoremstyle{plain}
\newtheorem{proposition}{Proposition}
\theoremstyle{remark}
\newtheorem{observation}{Observation}

\newcommand{\textquoted}[1]{\textquotedblleft #1\textquotedblright{}}
\graphicspath{{./}{./Figures/}{./Figures/unified-framework/}{./Figures/separation/}{./Figures/variable-order/}{./Figures/CDDDs/}}

\newcommand{\NOT}{\neg}
\newcommand{\AND}{\wedge}
\newcommand{\OR}{\vee}
\newcommand{\EQU}{\leftrightarrow}
\newcommand{\NOR}{\downarrow}
\newcommand{\DEC}{\diamond}

\newcommand{\G}{\mathcal{G}}

\newcommand{\BDD}{BDD}
\newcommand{\OBDD}[1]{OBDD$_{#1}$}
\newcommand{\ROBDD}[1]{ROBDD$_{#1}$}

\newcommand{\BDDC}[1]{BDD$[\AND_{#1}]$}
\newcommand{\OBDDC}[1]{O\BDDC{#1}}
\newcommand{\ROBDDC}[1]{R\OBDDC{#1}}

\newcommand{\MLDD}{MLDD}
\newcommand{\AOBDD}[1]{AOBDD$_{#1}$}

\newcommand{\OBDDL}[1]{OBDD-$L_{#1}$}
\newcommand{\ROBDDL}[1]{ROBDD-$L_{#1}$}

\newcommand{\Angle}[1]{\langle #1 \rangle}

\DontPrintSemicolon
\SetAlFnt{\scriptsize}
\SetKwBlock{KwFunc}{function}{end}

\newcommand{\EnsureFinest}{\textsc{Finest}}
\newcommand{\ExtractLeaf}{\textsc{ExtractLeaf}}
\newcommand{\ExtractPart}{\textsc{ExtractPart}}
\newcommand{\ExtractShare}{\textsc{ExtractShare}}

\newcommand{\Decompose}{\textsc{Decompose}}

\newcommand{\ConvertDown}{\textsc{ConvertDown}}

\newcommand{\oCO}{{CO}}
\newcommand{\oVA}{{VA}}
\newcommand{\oCE}{{CE}}
\newcommand{\oIM}{{IM}}
\newcommand{\oEQ}{{EQ}}
\newcommand{\oSE}{{SE}}
\newcommand{\oCT}{{CT}}
\newcommand{\oME}{{ME}}
\newcommand{\oCD}{{CD}}
\newcommand{\oSFO}{{SFO}}
\newcommand{\oFO}{{FO}}
\newcommand{\oABC}{{\AND BC}}
\newcommand{\oAC}{{\AND C}}
\newcommand{\oOBC}{{\OR BC}}
\newcommand{\oOC}{{\OR C}}
\newcommand{\oNC}{{\NOT C}}

\newcommand{\SE}{\textbf{SE}}

\newcommand{\CD}{\textbf{CD}}
\newcommand{\SFO}{\textbf{SFO}}
\newcommand{\FO}{\textbf{FO}}
\newcommand{\ABC}{$\mathbf{\AND}$\textbf{BC}}
\newcommand{\AC}{$\mathbf{\AND}$\textbf{C}}
\newcommand{\OBC}{$\mathbf{\OR}$\textbf{BC}}
\newcommand{\OC}{$\mathbf{\OR}$\textbf{C}}
\newcommand{\NC}{$\mathbf{\NOT}$\textbf{C}}

\begin{document}
%
\title{Augmenting Ordered Binary Decision Diagrams with Conjunctive Decomposition}
\author{Yong Lai$^{1,2,\ast}$, Dayou Liu$^{1,2}$, Minghao Yin$^{2,3}$\\
$^1$College of Computer Science and Technology, Jilin University, Changchun 130012, P.R. China\\
$^2$Key Laboratory of Symbolic Computation and Knowledge Engineering of Ministry of Education, \\Changchun 130012, P.R. China\\
$^3$College of Computer Science and Information Technology, Northeast Normal University, Changchun, P. R. China, 130117\\
laiy@jlu.edu.cn; liudy@jlu.edu.cn; ymh@nenu.edu.cn
}
\maketitle
\begin{abstract}
   This paper augments \OBDD{} with conjunctive decomposition to propose a generalization called \OBDDC{}. By imposing reducedness and the finest $\AND$-decomposition bounded by integer $i$ ($\AND_{\widehat{i}}$-decomposition) on \OBDDC{}, we identify a family of canonical languages called \ROBDDC{\widehat{i}}, where \ROBDDC{\widehat{0}} is equivalent to \ROBDD{}. We show that the succinctness of \ROBDDC{\widehat{i}} is strictly increasing when $i$ increases. We introduce a new time-efficiency criterion called rapidity which reflects that exponential operations may be preferable if the language can be exponentially more succinct, and show that the rapidity of each operation on \ROBDDC{\widehat{i}} is increasing when $i$ increases; particularly, the rapidity of some operations (e.g., conjoining) is strictly increasing. Finally, our empirical results show that: a) the size of \ROBDDC{\widehat{i}} is normally not larger than that of its equivalent \ROBDDC{\widehat{i+1}}; b) conjoining two \ROBDDC{\widehat{1}}s is more efficient than conjoining two \ROBDDC{\widehat{0}}s in most cases, where the former is NP-hard but the latter is in P; and c) the space-efficiency of \ROBDDC{\widehat{\infty}} is comparable with that of d-DNNF and that of another canonical generalization of \ROBDD{} called SDD.
\end{abstract}

\section{Introduction}
\label{sec:intro}

Knowledge Compilation (KC) is a key approach for dealing with the computational intractability in propositional reasoning \cite{Selman:Kautz:96,Darwiche:Marquis:02,Cadoli:Donini:97}. A core issue in the KC community is to identify target languages and then to evaluate them according to their properties. This paper focuses on three key properties: the canonicity of results of compiling knowledge bases into the language, the space-efficiency of storing compiled results, and the time-efficiency of operating compiled results. \cite{Darwiche:Marquis:02} proposed a KC map to characterize space-time efficiency by succinctness and tractability, where succinctness refers to the polysize transformation between languages, and tractability refers to the set of polytime operations a language supports. For an application, the KC map argues that one should first locate the necessary operations, and then choose the most succinct language that supports these operations in polytime.

Ordered Binary Decision Diagram (\OBDD{}) is one of the most influential KC languages in the literature \cite{Bryant:86}, due to twofold main theoretical advantages. First, its subset Reduced \OBDD{} (\ROBDD{}) is a canonical representation. Second, \ROBDD{} is one of the most tractable target languages which supports all the query operations and many transformation operations (e.g., conjoining) mentioned in the KC map in polytime. Despite its current success, a well-known problem with \OBDD{} is its weak succinctness, which reflects the explosion in size for many types of knowledge bases. Therefore, \cite{KAIS12} generalized \OBDD{} by associating some implied literals with each non-false vertex to propose a more succinct language called \OBDD{} with implied literals (\OBDDL{}). They showed that \OBDDL{} maintains both advantages of \OBDD{}. First, its subset ROBDD with implied literals as many as possible (\ROBDDL{\infty}) is also canonical. Second, given each operation ROBDD supports in polytime, \ROBDDL{\infty} also supports it in polytime in the sizes of the equivalent ROBDDs.

In order to further mitigate the size explosion problem of \ROBDD{} without loss of its theoretical advantages, we generalize \OBDDL{} by augmenting \OBDD{} with conjunctive decomposition to propose a language called \OBDDC{}. We then introduce a special type of $\AND$-decomposition called finest $\AND_{i}$-decomposition bounded by integer $i$ ($\AND_{\widehat{i}}$-decomposition), and impose reducedness and $\AND_{\widehat{i}}$-decomposition on \OBDDC{} to identify a family of canonical languages called \ROBDDC{\widehat{i}}. In particular, \ROBDDC{\widehat{0}} and \ROBDDC{\widehat{1}} are respectively equivalent to \ROBDD{} and \ROBDDL{\infty}. We show that the succinctness of \ROBDDC{\widehat{i}} is strictly stronger than \ROBDDC{\widehat{j}} if $i > j$. Our empirical results verify this property and also show that the space-efficiency of \ROBDDC{\widehat{\infty}} is comparable with that of deterministic Decomposable Negation Normal Form (d-DNNF, a superset of \OBDDC{}) \cite{Darwiche:01b} and that of another canonical subset called Sentential Decision Diagram (SDD) \cite{Darwiche:11} in d-DNNF.

We evaluate the tractability of \ROBDDC{\widehat{i}} and show that \ROBDDC{\widehat{i}} ($i > 0$) does not satisfy \SE{} (resp. \SFO{}, \ABC{} and \OBC{}) unless $\text{P} = \text{NP}$. According to the viewpoint of KC map, the applications which need the operation $OP$ corresponding to \SE{} (resp. \SFO{}, \ABC{} and \OBC{}) will prefer to \ROBDDC{\widehat{0}} than \ROBDDC{\widehat{1}}. In fact, the latter is strictly more succinct than the former, and also supports $OP$ in polytime in the sizes of the equivalent formulas in the former \cite{KAIS12}. In order to fix this \textquoted{bug}, we propose an additional time-efficiency evaluation criterion called rapidity which reflects an increase of at most polynomial multiples of time cost of an operation. We show that each operation on \ROBDDC{\widehat{i}} is at least as rapid as that on \ROBDDC{\widehat{j}} if $i \ge j$. In particular, some operations (e.g., conjoining) on \ROBDDC{\widehat{i}} are strictly more rapid than those on \ROBDDC{\widehat{j}} if $i > j$. Our empirical results verify that conjoining two \ROBDDC{\widehat{1}}s is more efficient than conjoining two \ROBDDC{\widehat{0}}s in most cases, where the former is NP-hard but the latter is in P.

\section{Basic Concepts}
\label{sec:decom}

We denote a propositional variable by $x$, and a denumerable variable set by $PV$. A formula $\varphi$ is constructed from constants $true$, $false$ and variables using negation operator $\NOT$, conjunction operator $\AND$ and disjunction operator $\OR$, and we denote by $Vars(\varphi)$ the set of its variables and by $PI(\varphi)$ the set of its prime implicates. The \emph{conditioning} of $\varphi$ on assignment $\omega$ ($ \varphi |_\omega$) is the formula obtained by replacing each appearance of $x$ in $\varphi$ by $true$ ($false$) if $x = true \text{ (}false\text{)} \in \omega$. $\varphi$ \emph{depends} on a variable $x$ iff $\varphi|_{x = false} \not\equiv \varphi|_{x = true}$. $\varphi$ is \emph{redundant} iff it does not depend on some $x \in Vars(\varphi)$. $\varphi$ is \emph{trivial} iff it depends on no variable.

\begin{definition}[$\AND$-decomposition]\label{def:decom}
A formula set $\Psi$ is a \emph{$\AND$-decomposition} of $\varphi$, iff $\varphi \equiv \bigwedge_{\psi \in \Psi} \psi$ and $\{Vars(\psi): \psi \in \Psi\}$ partitions $Vars(\varphi)$. A decomposition $\Psi$ is \emph{finer} than another $\Psi'$ iff $\{Vars(\psi):\psi \in \Psi\}$ is a refinement of $\{Vars(\psi):\psi \in \Psi'\}$; and $\Psi$ is \emph{strict} iff $|\Psi| > 1$.
\end{definition}

Let $\varphi$ be a non-trivial formula. If $\varphi$ is irredundant and $\{\psi_1, \ldots, \psi_m\}$ is its $\AND$-decomposition, $PI(\varphi) = PI(\psi_1) \cup \cdots \cup PI(\psi_m)$. If $\varphi$ does not depend on $x \in Vars(\varphi)$ and $\Psi$ is a $\AND$-decomposition of $\varphi|_{x = true}$, we can get a strict $\AND$-decomposition of $\varphi$ by adding $\NOT x \OR x$ to $\Psi$. Therefore,

\begin{proposition}\label{prop:uniq-dec}
From the viewpoint of equivalence, each non-trivial formula $\varphi$ has exactly one finest $\AND$-decomposition.
\end{proposition}

\begin{definition}[$\AND_i$-decomposition]\label{def:dec:bounded}
A $\AND$-decomposition $\Psi$ is \emph{bounded} by an integer $0 \leq i \leq \infty$ ($\AND_{i}$-decomposition) iff there exists at most one formula $\psi \in \Psi$ satisfying $|Vars(\psi)| > i$.
\end{definition}

Given a $\AND$-decomposition $\Psi$, we can get an equivalent $\AND_{i}$-decomposition by conjoining the formulas in $\Psi$ which has more than $i$ variables. According to Proposition \ref{prop:uniq-dec}, we have:

\begin{proposition}\label{prop:uniq-dec-spe}
For any non-trivial formula $\varphi$ and integer $0 \leq i \leq \infty$, $\varphi$ has exactly one finest $\AND_{i}$-decomposition from the viewpoint of equivalence.
\end{proposition}

Hereafter the finest $\AND_{i}$-decomposition is denoted by $\AND_{\widehat{i}}$-decomposition.

\section{\BDDC{} and Its Subsets}
\label{sec:defin}

In this section, we define \emph{binary decision diagram with conjunctive decomposition} (\BDDC{}) and some of its subsets.

\begin{definition}[\BDDC{}]\label{def:BDDC}
A \BDDC{} is a rooted directed acyclic graph. Each vertex $v$ is labeled with a symbol $sym(v)$: if $v$ is a leaf, $sym(v) = \bot/\top$; otherwise, $sym(v) = \AND$ (\emph{decomposition vertex}) or $sym(v) \in PV$ (\emph{decision vertex}).
Each internal vertex $v$ has a set of children $Ch(v)$; for a decision vertex, $Ch(v) = \{lo(v), hi(v)\}$, where $lo(v)$ and $hi(v)$ are called \emph{low} and \emph{high} children and connected by dashed and solid edges, respectively. Each vertex represents the following formula:
\begin{equation*}
\vartheta(v) = \begin{cases}
false/true & sym(v) = \bot/\top;\\
\bigwedge_{w \in Ch(v)}\vartheta(w) & {sym(v) = \AND;}  \\
\vartheta(lo(v)) \DEC_{sym(v)} \vartheta(hi(v)) & {\text{otherwise.}}
\end{cases}
\end{equation*}
where $\{\vartheta(w): w \in Ch(v)\}$ is a strict $\AND$-decomposition of $\vartheta(v)$ if $sym(v) = \AND$, and $\varphi \DEC_{x} \psi = (\NOT x \AND \varphi) \OR (x \AND \psi)$. The formula represented by the \BDDC{} is defined as the one represented by its root.
\end{definition}

Hereafter we denote a leaf vertex by $\Angle{\bot/\top}$, a decomposition vertex ($\AND$-vertex for short) by $\Angle{\AND, Ch(v)}$, and a decision vertex ($\DEC$-vertex for short) by $\langle sym(v), lo(v), hi(v) \rangle$. We abuse $\Angle{\AND, \{w\} }$ to denote $w$, $\Angle{\AND, \emptyset}$ to denote $\Angle{\top}$, $\Angle{\AND, \{\Angle{\top}\} \cup V}$ to denote $\Angle{\AND, V}$, and $\Angle{\AND, \{\Angle{\bot}\} \cup V}$ to denote $\Angle{\bot}$.
Given a \BDDC{} $\G$, $|\G|$ denotes the size of $\G$ defined as the number of its edges. In addition, we use $\G_v$ to denote the \BDDC{} rooted at $v$, and occasionally abuse $v$ to denote $\vartheta(v)$.
Now we define the subsets of \BDDC{}:

\begin{definition}[subsets of \BDDC{}]\label{def:BDDC:frag}
A \BDDC{} is \emph{ordered} over a linear order of variables $\prec$ over $PV$ (\OBDDC{}) iff each $\DEC$-vertex $u$ and its $\DEC$-descendant $v$ satisfy $sym(u) \prec sym(v)$. 
An \OBDDC{} is \emph{reduced} (\ROBDDC{}), iff no two vertices are identical (having the same symbol and children) and no $\DEC$-vertex has two identical children. 
An \OBDDC{} is \emph{$\AND_{i}$-decomposable} (\OBDDC{i}) iff each $\AND$-vertex is a $\AND_{i}$-decomposition. 
An \ROBDDC{}{}{} is \emph{$\AND_{\widehat{i}}$-decomposable} (\ROBDDC{\widehat{i}}{}{}), iff each $\AND$-vertex is a $\AND_{\widehat{i}}$-decomposition and the $\AND_{\widehat{i}}$-decomposition of each $\DEC$-vertex $v$ is $\{v\}$.\footnote{Each internal vertex in \ROBDDC{} is non-trivial.}
\end{definition}

\begin{figure}[ht]
  \centering
  \subfloat[]{\label{fig:example:a}
    \centering
    \begin{minipage}[c]{0.35\linewidth}
        \centering
        \includegraphics[width = \textwidth]{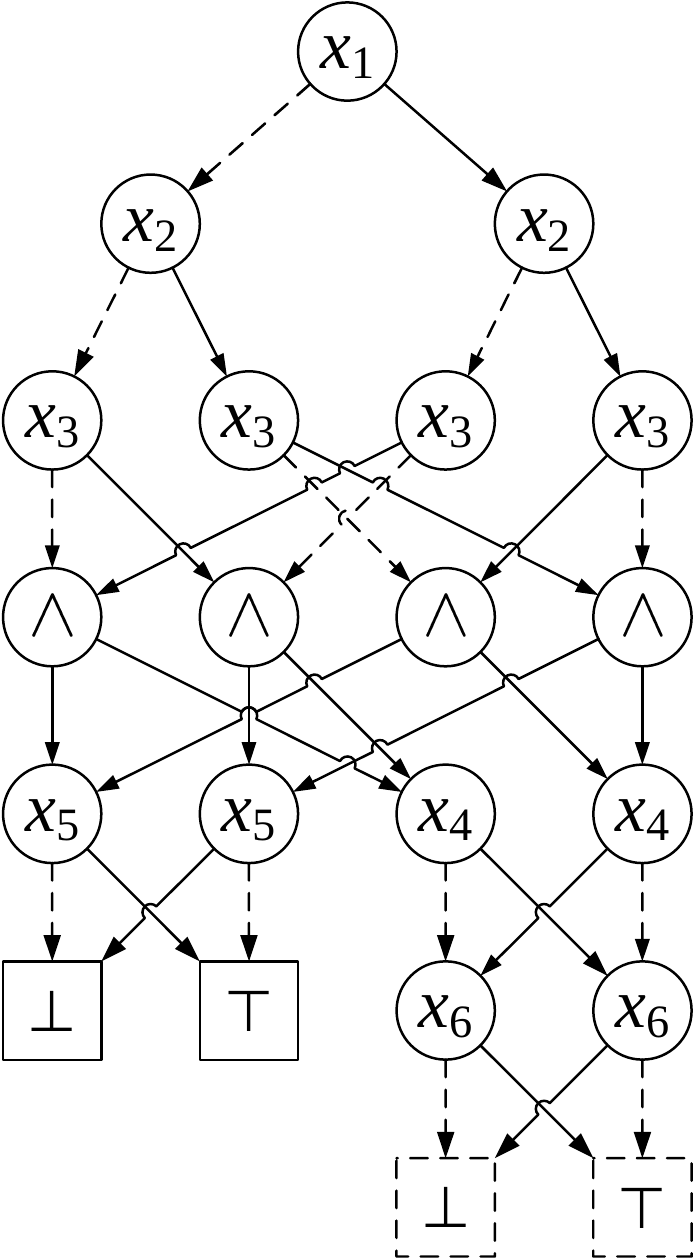}
    \end{minipage}
  }
  \begin{minipage}[c]{0.1\linewidth}
    \hspace{2mm}
  \end{minipage}
  \subfloat[]{\label{fig:example:b}
    \centering
    \begin{minipage}[c]{0.35\linewidth}
        \centering
        \includegraphics[width = \textwidth]{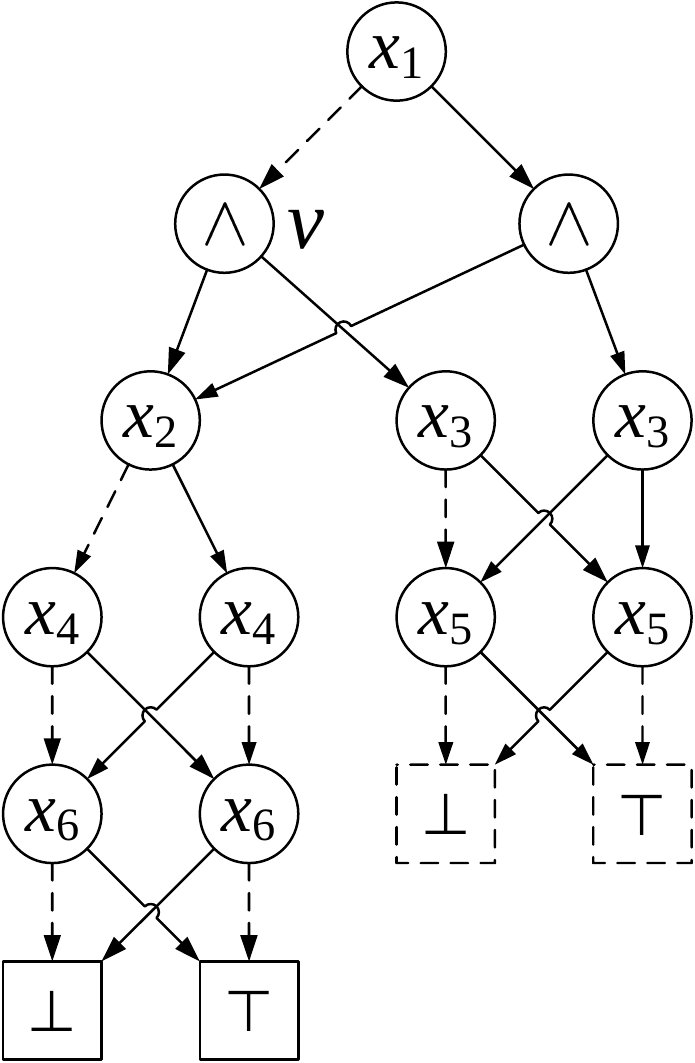}
    \end{minipage}
  }
  \caption{An \ROBDDC{\widehat{1}} (a) and an \ROBDDC{\widehat{2}} (b)}\label{fig:example}
\end{figure}

For any two \OBDDC{}s, unless otherwise stated, hereafter assume that they are over the same variable order and $x_k$ is the $k$th variable. In the following, we mainly focus on \ROBDDC{\widehat{i}}, and analyze its canonicity and space-time efficiency. Obviously, \ROBDD{} is equivalent to \ROBDDC{\widehat{0}}. In addition, since a \BDD{} vertex labelled with a set of implied literals in \cite{KAIS12} can be seen as a $\AND_1$-vertex, it is easy to show \ROBDDL{\infty} is equivalent to \ROBDDC{\widehat{1}}. Figures \ref{fig:example:a} and \ref{fig:example:b} respectively depict an \ROBDDC{\widehat{1}} and an \ROBDDC{\widehat{2}} representing $\varphi = (x_1 \EQU x_3 \EQU x_5) \AND (x_2 \EQU x_4 \EQU x_6)$. Note that for simplicity, we draw multiple copies of vertices, denoted by dashed boxes, but they represent the same vertex. Figure \ref{fig:example:b} is not an \OBDDC{1} since vertex $v$ is not bounded by one. If we extend $\varphi$ to the following formula, the number of vertices labelled with $x_{1+n}$ in \ROBDDC{\widehat{j}} will equal $2^n$, while the number of vertices in \ROBDDC{\widehat{i}} ($i > j$) will be $(2j+5)n$. That is, the size of \ROBDDC{\widehat{j}} representing Eq. (\ref{eq:equ-and}) is exponential in $n$, while the size of the corresponding \ROBDDC{\widehat{i}} is only linear in $n$.
\begin{equation}\label{eq:equ-and}
\bigwedge_{1 \le k \le n} x_{k + 0 \cdot n} \leftrightarrow \cdots \leftrightarrow x_{k + (j+1) \cdot n}
\end{equation}

We close this section by pointing out that \ROBDDC{\widehat{i}} is canonical and complete. The canonicity is immediately from the uniqueness of $\AND_{\widehat{i}}$-decomposition. The completeness is also easily understood, since we can transform \ROBDD{} into \ROBDDC{\widehat{i}} (see the next section).

\begin{proposition}\label{prop:canon}
Given a formula, there is exactly one \ROBDDC{\widehat{i}} to represent it.
\end{proposition}

\section{Space-Efficiency Analysis}
\label{sec:space}

We analyze the space-efficiency in terms of succinctness \cite{Gogic:etal:95,Darwiche:Marquis:02}. The succinctness results is given as follows, where $\text{L} \le_s \text{L}'$ denotes that $\text{L}$ is not more succinct than $\text{L}'$. Due to space limit we just briefly explain them, from two aspects.
\begin{proposition}\label{prop:succinct:1}
\ROBDDC{\widehat{i}} $\le_s$ \ROBDDC{\widehat{j}} iff $i \le j$.
\end{proposition}

First, we show the direction from right to left by proposing an algorithm called \Decompose{} (in Algorithm \ref{alg:Decompose}) which can transform an \OBDDC{i} into the equivalent \ROBDDC{\widehat{i}} in polytime. Note that each \ROBDDC{\widehat{i}} is an \OBDDC{j}. \Decompose{} immediately provides a compiling method of \ROBDDC{\widehat{i}}, that is, transforming \ROBDDC{\widehat{0}} (i.e., \ROBDD{}) into \ROBDDC{\widehat{i}}. Some functions used in \Decompose{} are explained as follows:
\begin{itemize}
\item \EnsureFinest($u'$): While there exists some $v \in Ch(u')$ with $sym(u') = sym(v) = \AND$, we repeat removing $v$ from $Ch(u')$ and then adding all children of $v$ to $Ch(u')$.
\item \ExtractLeaf($u'$): The input $u'$ of this function, as well as the next two functions, is a $\DEC$-vertex whose children are \ROBDDC{\widehat{i}} vertices; we employ these functions to get an \ROBDDC{\widehat{i}} vertex equivalent to $u'$. \ExtractLeaf{} handles the case where $\Angle{\bot} \in Ch(u')$ and $|Vars(u')| > 1$. Without loss of generality, assume $lo(u') = \Angle{\bot}$. If $i = 0$, $u'$ is already an \ROBDDC{\widehat{i}} vertex. Otherwise, let $v =\Angle{\AND, \{\Angle{sym(u'), \Angle{\bot}, \Angle{\top}}, hi(u') \}} \equiv u'$. Then we call $\EnsureFinest(v)$ to get an equivalent \ROBDDC{\widehat{i}} vertex.
\item \ExtractPart($u'$): This function handles the case where one child of $u'$ is a part of the other. That is, $Ch(u') = \{v_1, v_2\}$ satisfies $v_1 \in Ch(v_2)$ and  $sym(v_2) = \AND$. Without loss of generality, assume $v_1 = lo(u')$. Let $v = \Angle{sym(u'), \Angle{\top}, \Angle{\AND, Ch(v_2) \setminus \{v_1\} }}$. Then $u'' = \Angle{\AND, \{v_1, v\}} \equiv u'$. If $i < |Vars(v_1)|$ and $i < |Vars(v)|$, then $u'$ is already an \ROBDDC{\widehat{i}} vertex. Otherwise, $u''$ is an \ROBDDC{\widehat{i}} vertex.
\item \ExtractShare($u'$): This function handles the case where $lo(u')$ and $hi(u')$ share some children. That is, $sym(lo(u')) = sym(hi(u')) = \AND$ and $V = Ch(lo(u')) \cap Ch(hi(u')) \not= \emptyset$. If $V = Ch(lo(u'))$, we return $lo(u')$. Otherwise, if there exists some $v \in V$ with $|Vars(v)| > i$ and $|Vars(u') \setminus Vars(V)| > i$, we remove $v$ from $V$. If $V = \emptyset$, $u'$ is already an \ROBDDC{\widehat{i}} vertex. Otherwise, let $v = \Angle{sym(u'), lo(u') \setminus V, hi(u') \setminus V}$, and then $\Angle{\AND, V \cup \{v\}}$ is an \ROBDDC{\widehat{i}} vertex equivalent to $u'$.
\end{itemize}

\begin{algorithm}[htbp]
\caption{\Decompose($u$)} \label{alg:Decompose}
\KwIn{an \OBDDC{i} vertex $u$}
\KwOut{the \ROBDDC{\widehat{i}} vertex which is equivalent to $u$}
\lIf {$H(u) \not= nil$} {\KwRet $H(u)$}\;
\lIf {$u$ is a leaf vertex} {$u' \leftarrow u$}\;
\Else {
    $u' \leftarrow \Angle{sym(u), \{\Decompose(v) : v \in Ch(u)\}}$\;
    \uIf {$u'$ is a $\DEC$-vertex} {
        \uIf {$\Angle{\bot} \in Ch(u')$ and $|Vars(u')| > 1$} {
            $u' \leftarrow \ExtractLeaf(u')$
        }
        \uElseIf {one child of $u$ is a part of the other} {
            $u' \leftarrow \ExtractPart(u')$
        }
        \uElseIf {the children of $u$ share some children} {
            $u' \leftarrow \ExtractShare(u')$
        }
        \lElseIf {$lo(u') = hi(u')$} {$u' \leftarrow lo(u')$}\;
     }
    \lElse {$u' \leftarrow \EnsureFinest(u)$}\;
}
\lIf {a previous vertex $u''$ identical with $u'$ appears} {$H(u) \leftarrow u''$}\;
\lElse {$H(u) \leftarrow u'$}\;
\KwRet $H(u)$
\end{algorithm}

Second, we show the direction from left to right by counterexample: If $i > j$,  Eq. (\ref{eq:equ-and})
can be represented by an \ROBDDC{\widehat{i}} in linear size, but the size of the equivalent \ROBDDC{\widehat{j}} is exponential in $n$.

\section{Time-Efficiency Analysis}
\label{sec:tract}

We analyze the time-efficiency of operating \ROBDDC{\widehat{i}} in terms of tractability \cite{Darwiche:Marquis:02} and a new perspective. First we present the operations mentioned in this paper:

\begin{definition}[operation]\label{def:operation}
An operation $OP$ is a relation between $\Delta_p \times \Delta_s$ and $\Gamma$, where $\Delta_p$ denotes the primary information of $OP$ which is a set of sequences of formulas, $\Delta_s$ denotes the supplementary information customized for $OP$, and $\Gamma$ is the set of outputs of $OP$. $OP$ on language L, denoted by $OP(\text{L})$, is the subset $\{(((\varphi_1, \ldots, \varphi_n), \alpha), \beta) \in OP: \varphi_i \in \text{L} \text{ for }1 \le i \le n \text{ and }\beta \in \text{L} \text{ if it is a formula}\}$.
\end{definition}

Hereafter, we abbreviate $(((\varphi_1, \ldots, \varphi_n), \alpha), \beta) \in OP$ as $(\varphi_1, \ldots, \varphi_n, \alpha, \beta) \in OP$. According to the above definition, we can easily formalize the query operations ($\oCO$, $\oVA$, $\oCE$, $\oIM$, $\oEQ$, $\oSE$, $\oCT$ and $\oME$) and transformation operations ($\oCD$, $\oSFO$, $\oFO$, $\oABC$, $\oAC$, $\oOBC$, $\oOC$ and $\oNC$) mentioned in the KC map. We say an algorithm $\textsc{Alg}$ performs operation $OP(\text{L})$, iff for each $(\varphi_1, \ldots, \varphi_n, \alpha, \beta) \in OP(\text{L})$, $(\varphi_1, \ldots, \varphi_n, \alpha, \textsc{Alg}(\varphi_1, \ldots, \varphi_n, \alpha)) \in OP(\text{L})$.

\subsection{Tractability Evaluation}
\label{sec:tract:KC}

As the KC map, we say language L satisfies $\textbf{OP}$ iff there exists some polytime algorithm performing $OP(\text{L})$. The tractability results are shown in Table \ref{tab:tract}, and due to space limit we will only discuss proofs of the less obvious ones.

\begin{table}[!htbp]
\centering
\scriptsize
\caption{The polytime queries and transformations, where $\surd$ means \textquoted{satisfies}, $\bullet$ means \textquoted{does not satisfy}, and $\circ$ means \textquoted{does not satisfy unless P $=$ NP}}\label{tab:tract}
\vspace{-7pt}
\renewcommand{\arraystretch}{1.3}
\begin{tabularx}{\linewidth}{|@{\hspace{4.5pt}}>{\centering\arraybackslash}c@{\hspace{4.5pt}} |*{8}{@{\hspace{2.5pt}}>{\centering\arraybackslash}X@{\hspace{2.5pt}}|}}\hline
L                 & \textbf{CO} & \textbf{VA} & \textbf{CE} & \textbf{IM} & \textbf{EQ} & \textbf{SE} & \textbf{CT} & \textbf{ME}  \\\hline
\ROBDDC{\widehat{0}} & $\surd$ & $\surd$ & $\surd$ & $\surd$ & $\surd$ & $\surd$ & $\surd$ & $\surd$  \\\hline
\ROBDDC{\widehat{i}} ($i>0$) & $\surd$ & $\surd$ & $\surd$ & $\surd$ & $\surd$ & $\circ$ & $\surd$ & $\surd$  \\\hline
L                 & \CD{} & \FO{} & \SFO{} & \AC{} & \ABC{} & \OC{} & \OBC{} & \NC{}  \\\hline
\ROBDDC{\widehat{0}} & $\surd$ & $\bullet$ & $\surd$   & $\bullet$ & $\surd$ & $\bullet$ & $\surd$ & $\surd$  \\\hline
\ROBDDC{\widehat{i}} ($i>0$) & $\surd$ & $\circ$ & $\circ$ & $\circ$ & $\circ$ & $\circ$ & $\circ$ & $\bullet$  \\\hline
\end{tabularx}
\end{table}


Since \ROBDDC{\widehat{i}} is a subset of d-DNNF, it supports each query operation which is tractable for d-DNNF, in polytime. According to the following observation, \ROBDDC{\widehat{i}} ($i > 0$) does not satisfy \SE{} unless P $=$ NP, which implies that \ROBDDC{\widehat{i}} does not satisfy \ABC{} (resp. \AC, \OBC, \OC, \SFO{} and \FO{}) unless P $=$ NP. Observation \ref{obs:imp:SE} can be proved by modifying the proof of Theorem 3.1 in \cite{Fortune:etal:78}, since both free BDDs in the proof can be replaced by \ROBDDC{\widehat{i}}.
\begin{observation}\label{obs:imp:SE}
Given any two \ROBDDC{\widehat{i}} ($i > 0$) vertices $u$ and $v$, the problem of deciding whether $u \models v$ holds is co-NP-complete.
\end{observation}

Given a \BDDC{} vertex $u$ and an assignment $\omega$, we can get a vertex $u'$ equivalent to $u|_\omega$ by replacing each $\Angle{x, v, v'}$ appearance in $\G_{u}$ with $\Angle{x, v, v}$ ($\Angle{x, v', v'}$) for each $x = false \text{ (}true \text{)} \in \omega$. We can call \Decompose{} to transform $u'$ into \ROBDDC{\widehat{i}} in polytime if $u$ is in \ROBDDC{\widehat{i}}. That is, \ROBDDC{\widehat{i}} satisfies \CD{}. If $u$ is in \ROBDDC{\widehat{i}}, we use $u \downharpoonleft_\omega$ to denote the \ROBDDC{\widehat{i}} vertex which is equivalent to $u |_\omega$. Finally, the \ROBDDC{\widehat{i}} ($i > 0$) $\G$ representing Eq. (\ref{eq:equ-and}) has a linear size, but the negation of $\G$ has an exponential size. That is, \ROBDDC{\widehat{i}} does not satisfy \NC{}.

\subsection{A New Perspective About Time-Efficiency}

Due to distinct succinctness, it is sometimes insufficient to compare the time-efficiency of two canonical languages by comparing their tractability. For example, according to the tractability results mentioned previously, \ROBDDC{\widehat{1}} does not satisfy \SE{} (resp. \SFO{}, \ABC{} and \OBC{}) unless $\text{P} = \text{NP}$. From the perspective of KC map, the applications which need the operation $OP \in \{\oSE{}, \oSFO{}, \oABC{}, \oOBC{}\}$ will prefer to \ROBDDC{\widehat{0}} than \ROBDDC{\widehat{1}}. In fact, the latter is strictly more succinct than the former, and also supports $OP$ in polytime in the sizes of the equivalent formulas in the former \cite{KAIS12}. In order to fix this \textquoted{bug}, we propose an additional time-efficiency evaluation criterion tailored for canonical languages.

\begin{definition}[rapidity]\label{def:rapidity}
Given an operation $OP$ and two canonical languages $\text{L}$ and $\text{L}'$, $OP(\text{L})$ is \emph{at least as rapid as} $OP(\text{L}')$ ($\text{L}' \le_{r}^{OP} \text{L}$), iff for each algorithm $\textsc{Alg}'$ performing $OP(\text{L}')$, there exists some polynomial $p$ and some algorithm $\textsc{Alg}$ performing $OP(\text{L})$ such that for every input $(\varphi_1, \ldots, \varphi_n, \alpha) \in OP(\text{L})$ and its equivalent input $(\varphi_1', \ldots, \varphi_n', \alpha) \in OP(\text{L}')$, $\textsc{Alg}$($\varphi_1, \ldots, \varphi_n, \alpha$) can be done in time $p(t + |\varphi_1'| + \cdots + |\varphi_n'| + |\alpha|)$, where $t$ is the running time of $\textsc{Alg}'$($\varphi_1', \ldots, \varphi_n', \alpha$).
\end{definition}

Note that the rapidity relation is reflexive and transitive. Let $OP$ be an operation, and $\text{L}$ and $\text{L}'$ be two canonical languages, where $\text{L} \le_r^{OP} \text{L}'$. Given each input $(\varphi_1, \ldots, \varphi_n, \alpha) \in OP(\text{L})$ and its equivalent input $(\varphi_1', \ldots, \varphi_n', \alpha) \in OP(\text{L}')$, time cost of performing $OP$ on $(\varphi_1, \ldots, \varphi_n, \alpha)$ increases at most polynomial times than that of performing $OP$ on $(\varphi_1', \ldots, \varphi_n', \alpha)$. In particular, if $\text{L}'$ supports $OP$ in polytime, then $\text{L}$ also supports $OP$ in polytime in sizes of the equivalent formulas in $\text{L}'$. Thus for applications needing canonical languages, we suggest that one first identify the set $\mathcal{OP}$ of necessary operations, second identify the set $\mathcal{L}$ of canonical languages meeting the tractability requirements, third add any canonical language $\text{L}$ satisfying $\exists \text{L}' \in \mathcal{L} \forall  OP \in \mathcal{OP}. \text{L} \le_r^{OP} \text{L}'$ to $\mathcal{L}$, and finally choose the most succinct language in $\mathcal{L}$. Now we present the rapidity results:

\begin{proposition}\label{prop:rapid}
\ROBDDC{\widehat{i}} $\le_r^{OP}$ \ROBDDC{\widehat{j}} if $i \le j$. In particular, for $OP \in \{\oCD, \oFO,\oSFO, \oAC,\oABC, $ $\oOBC, \oOC\}$, \ROBDDC{\widehat{i}} $\not\le_r^{OP}$ \ROBDDC{\widehat{j}} if $i > j$.
\end{proposition}

We emphasize an interesting observation here. It was mentioned that for $OP \in \{\oSE, \oSFO, \oABC, \oOBC\}$, $OP$(\ROBDDC{\widehat{0}}) can be performed in polytime but $OP$(\ROBDDC{\widehat{i}}) ($i>0$) cannot be performed in polytime unless $\text{P} = \text{NP}$. Therefore, if we only consider the tractability of $OP$, it may lead to the illusion that the time efficiency of performing $OP$(\ROBDDC{\widehat{i}}) is pessimistically lower than that of performing $OP$(\ROBDDC{\widehat{0}}). In fact, Proposition \ref{prop:rapid} shows that $OP$(\ROBDDC{\widehat{i}}) can also be performed in polytime in the sizes of equivalent \ROBDDC{\widehat{0}}s. That is, according to our new perspective, the applications which need $OP$ will prefer to \ROBDDC{\widehat{i}} than \ROBDDC{\widehat{0}}.

To explain the first conclusion in Proposition \ref{prop:rapid}, we first propose an algorithm called \ConvertDown{} (in Algorithms \ref{alg:ConvertDown}) to transform \ROBDDC{\widehat{j}} into \ROBDDC{\widehat{i}}. \ConvertDown{} terminates in polytime in the size of output, due to the facts that \ROBDDC{\widehat{i}} $\le_s$ \ROBDDC{\widehat{j}} and that \ROBDDC{\widehat{j}} is canonical and satisfies \CD{}. \ConvertDown{}, together with \Decompose{}, provide new methods to answer query and to perform transformation on \ROBDDC{\widehat{j}}. First, we call \ConvertDown{} to transform \ROBDDC{\widehat{j}}s into \ROBDDC{\widehat{i}}s. Next, we answer query using the outputs of the first step, or perform transformation on the outputs and then transform the result into \ROBDDC{\widehat{j}} by \Decompose{}. Since the time complexities of \Decompose{} and \ConvertDown{} are polynomial in the sizes of \ROBDDC{\widehat{i}}s, we know \ROBDDC{\widehat{i}} $\le_r^{OP}$ \ROBDDC{\widehat{j}}.

\begin{algorithm}[!htb]
\caption{\ConvertDown($u$)} \label{alg:ConvertDown}
\KwIn{an \ROBDDC{\widehat{j}} rooted at $u$}
\KwOut{the \ROBDDC{\widehat{i}} representing $\vartheta(u)$, where $i \le j$}
\lIf {$H(u) \not= nil$} {\KwRet $H(u)$}\;
\lIf {u is a leaf} {$H(u) \leftarrow u$}\;
\uElseIf {u is a $\DEC$-vertex} {
    $\ConvertDown(lo(v))$; $\ConvertDown(hi(v))$\;
    $H(u) \leftarrow \langle sym(u), H(lo(u)), H(hi(u)) \rangle$
}
\Else {
    $V \leftarrow \{v \in Ch(u): |Vars(v)| > i\}$\;
    \lIf {$|V| \leq 1$} {$H(u) \leftarrow u$}\;
    \Else {
        Let $v$ be $\Angle{sym(u), V}$ and $x$ be the least variable in $Vars(v)$\;
        $v' \leftarrow \ConvertDown(\Angle{x, v \downharpoonleft_{x = false}, v \downharpoonleft_{x = true}})$\;
        $H(u) \leftarrow \langle sym(u), (Ch(u) \setminus V) \cup \{v'\} \rangle$\;
    }
}
\KwRet $H(u)$
\end{algorithm}

Now we turn to explain the second conclusion in Proposition \ref{prop:rapid}. Due to the facts that $\textrm{L} \not\le_r^{\oSFO} \textrm{L}'$ iff $\textrm{L} \not\le_r^{\oOBC} \textrm{L}'$, $\textrm{L} \not\le_r^{\oSFO} \textrm{L}'$ implies $\textrm{L} \not\le_r^{\oFO} \textrm{L}'$, $\textrm{L} \not\le_r^{\oABC} \textrm{L}'$ implies $\textrm{L} \not\le_r^{\oAC} \textrm{L}'$, and $\textrm{L} \not\le_r^{\oOBC} \textrm{L}'$ implies $\textrm{L} \not\le_r^{\oOC} \textrm{L}'$, we just need to show the cases when $OP \in \{\oCD, \oABC, \oOBC\}$.

\ROBDDC{\widehat{i}} $\not\le_r^{\oCD}$ \ROBDDC{\widehat{j}}: Conditioning the \ROBDDC{\widehat{j}} representing  Eq. (\ref{eq:equ-and}) on $\{x_{n + 0 \cdot n} = true, \ldots,$ $ x_{n + (j+1) \cdot n} = true\}$ will introduce an exponential (in the size of the equivalent \ROBDDC{\widehat{i}}) number of new vertices.

\ROBDDC{\widehat{i}} $\not\le_r^{\oABC}$ \ROBDDC{\widehat{j}}: Consider the formulas $\bigwedge_{1 \le k \le n}x_{k + 0 \cdot n} \OR \NOT(x_{k + 1 \cdot n} \leftrightarrow \cdots \leftrightarrow x_{k + (j+1) \cdot n})$ and $\bigwedge_{1 \le k \le n}\NOT x_{k + 0 \cdot n} \OR (x_{k + 1 \cdot n} \leftrightarrow \cdots \leftrightarrow x_{k + (j+1) \cdot n})$. It is easy to design an algorithm to conjoin the \ROBDDC{\widehat{i}}s representing the above two formulas in linear time. However, the conjunction of two \ROBDDC{\widehat{j}}s representing the two formulas will generate the \ROBDDC{\widehat{j}} representing Eq. (\ref{eq:equ-and}) with an exponential number of new vertices.


\ROBDDC{\widehat{i}} $\not\le_r^{\oOBC}$ \ROBDDC{\widehat{j}}: By replacing the two formulas mentioned above with the following ones: $x_{1 + 0 \cdot n} \AND (x_{1 + 1 \cdot n} \leftrightarrow \cdots \leftrightarrow x_{1 + (j+1) \cdot n}) \AND \bigwedge_{2 \le k \le n} x_{k + 0 \cdot n} \leftrightarrow \cdots \leftrightarrow x_{k + (j+1) \cdot n}$ and $\NOT x_{1 + 0 \cdot n} \AND \NOT(x_{1 + 1 \cdot n} \leftrightarrow \cdots \leftrightarrow x_{1 + (j+1) \cdot n}) \AND \bigwedge_{2 \le k \le n} x_{k + 0 \cdot n} \leftrightarrow \cdots \leftrightarrow x_{k + (j+1) \cdot n}$, we can prove this conclusion in a similar way.

\section{Preliminary Experimental Results}
\label{sec:experiment}

In this section, we report some preliminary experimental results of \ROBDDC{\widehat{i}} ($0 \le i \le \infty$), to verify several previous theoretical properties. In our experiments about space-efficiency, each CNF formula was first compiled into \ROBDDC{\widehat{1}} by the \ROBDDL{\infty} compiler in \cite{KAIS12} under the min-fill heuristic, then the resulting \ROBDDC{\widehat{1}} was transformed into \ROBDDC{\widehat{\infty}} by \Decompose{}, and finally the resulting \ROBDDC{\widehat{\infty}} was transformed into \ROBDDC{\widehat{i}} by \ConvertDown{}. We also compared the conjoining efficiency of \ROBDDC{\widehat{0}} with that of \ROBDDC{\widehat{1}}. All experiments were conducted on a computer with a two-core 2.99GHz CPU and 3.4GB RAM.

\begin{figure*}[!htb]
  \centering
  \includegraphics[width = 0.48\linewidth]{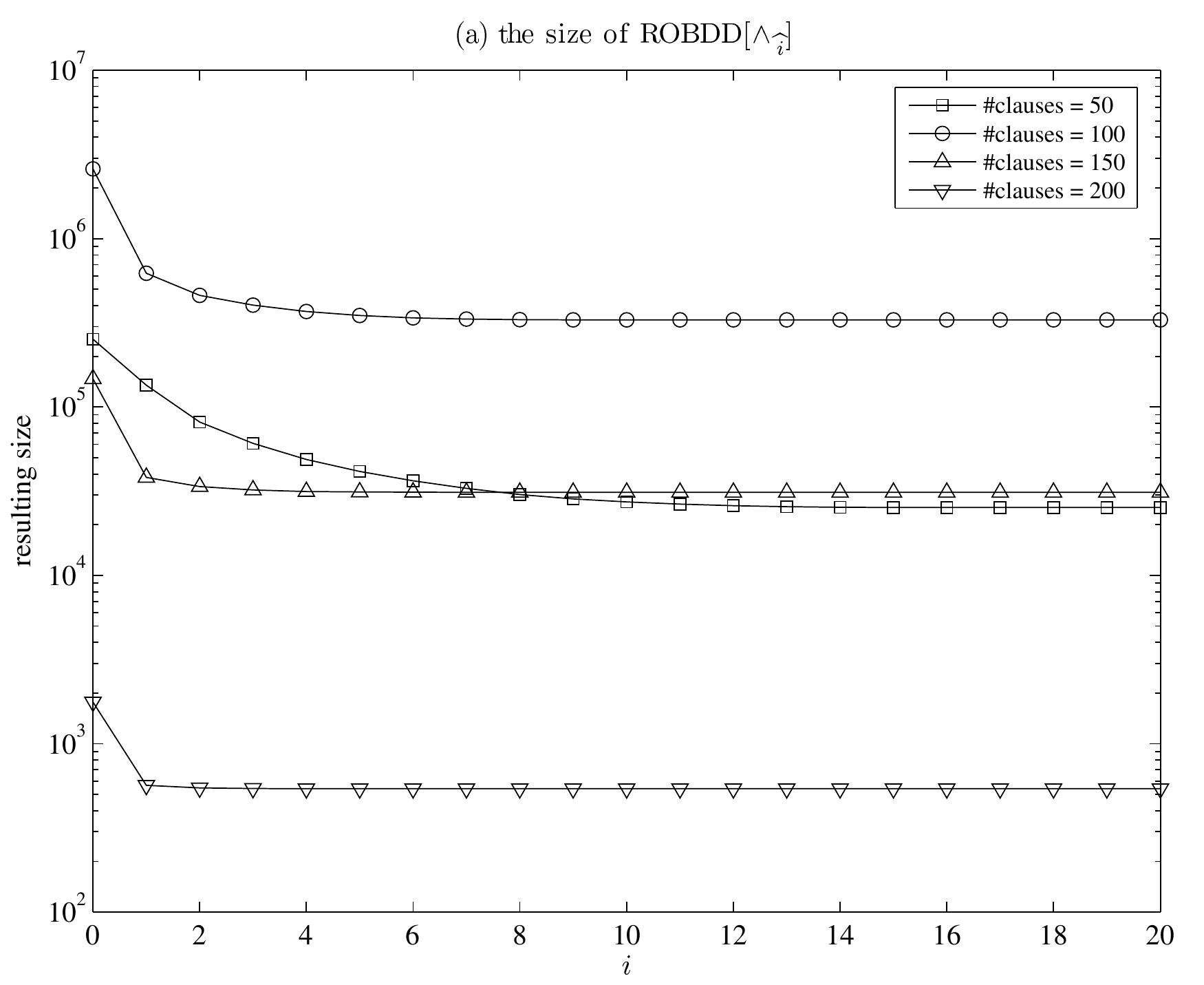}
  \includegraphics[width = 0.48\linewidth]{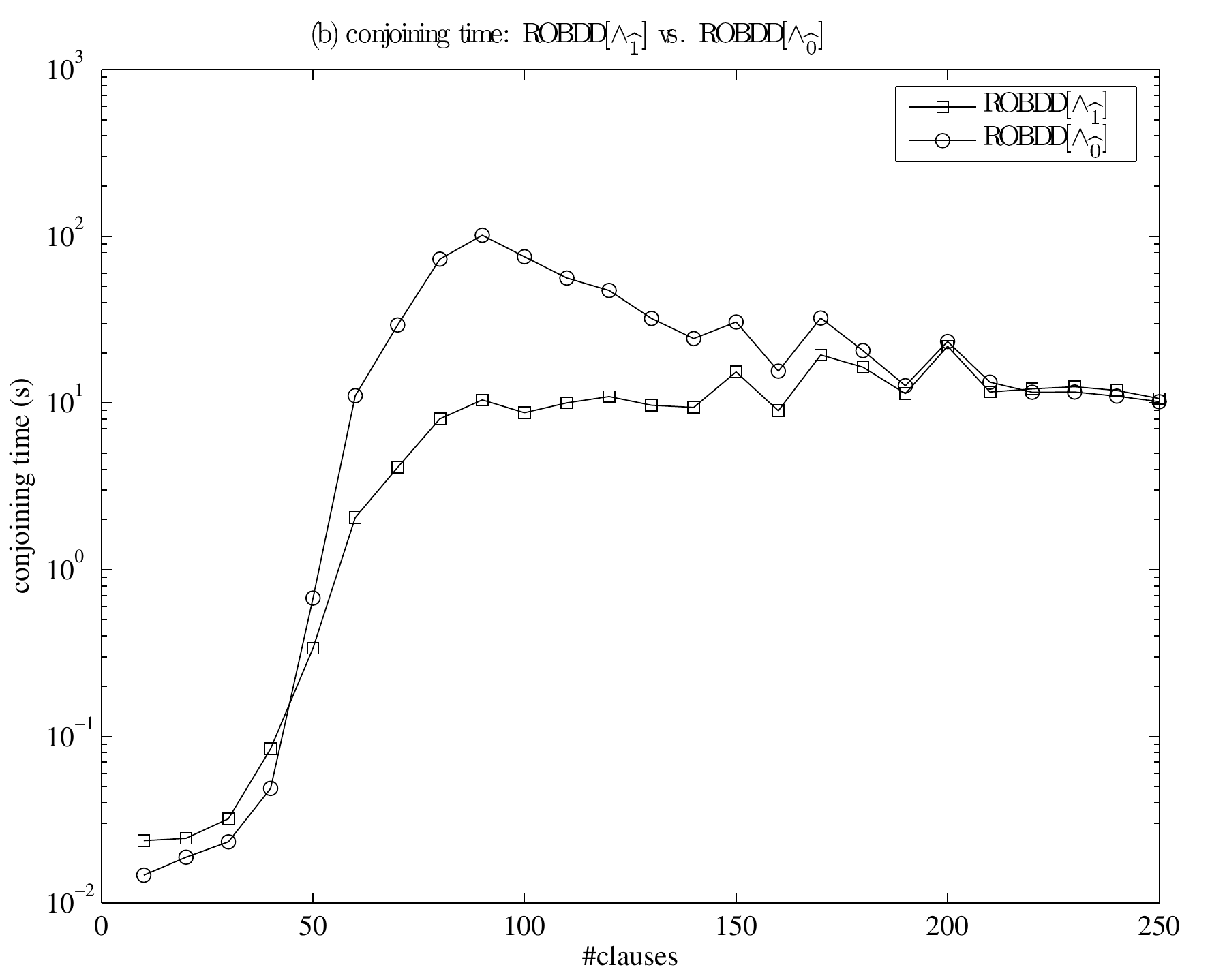}\\
  \includegraphics[width = 0.48\linewidth]{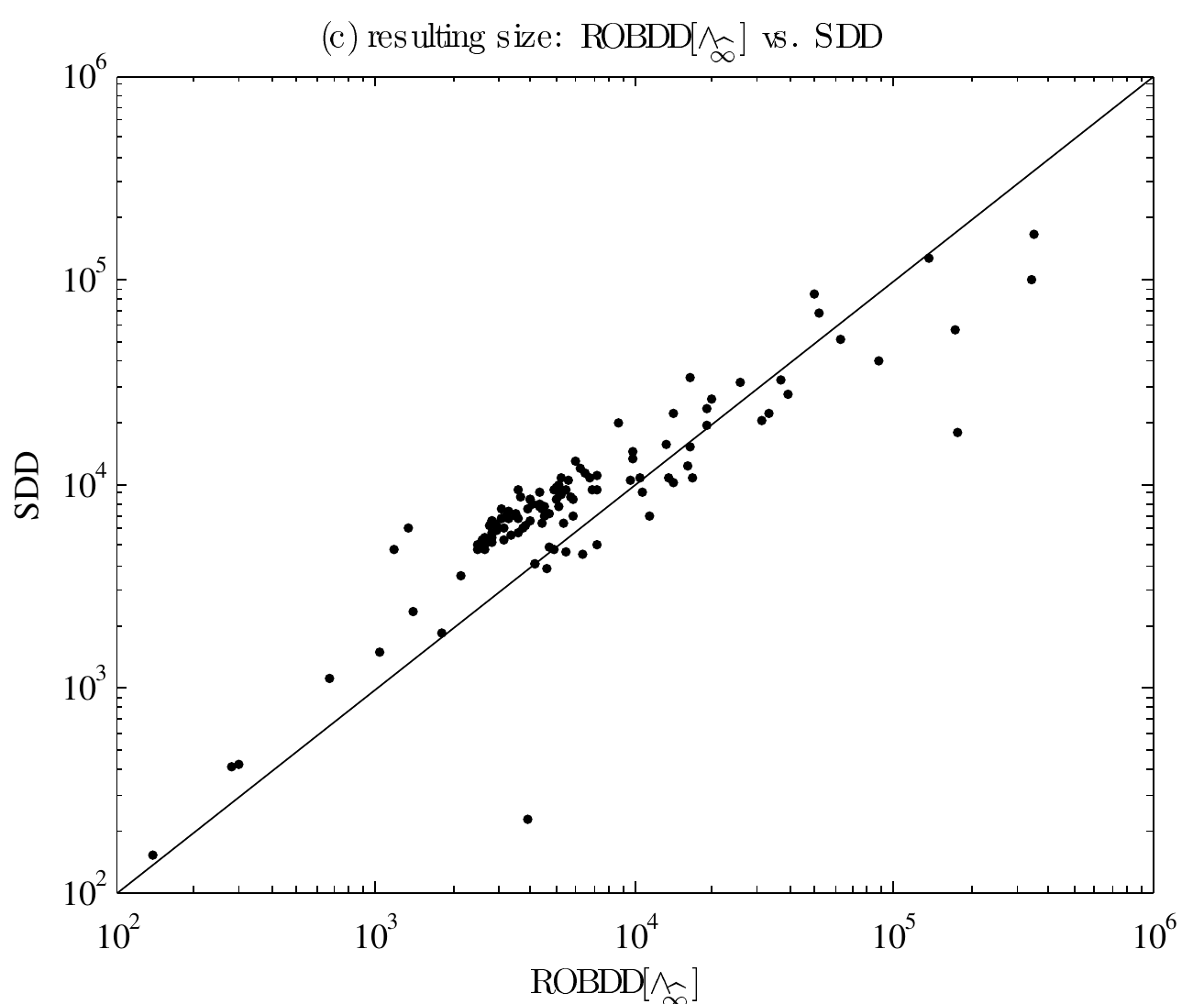}
  \includegraphics[width = 0.48\linewidth]{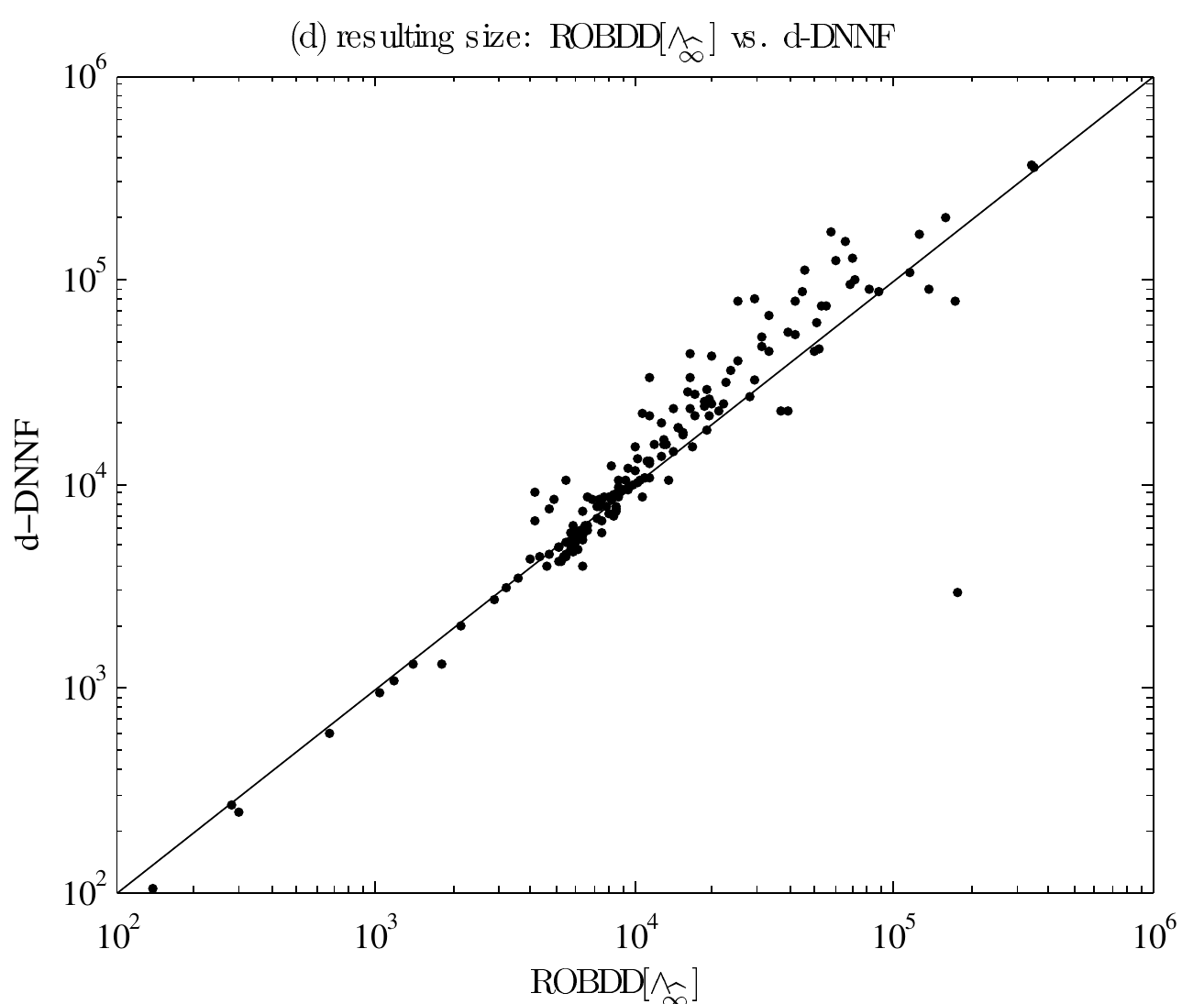}\\
  \caption{An empirical evaluation of the \ROBDDC{\widehat{i}} ($0 \le i \le \infty$) size and the conjoining time of \ROBDDC{\widehat{0}} and \ROBDDC{\widehat{1}}, where (a-b) are on random 3-CNF instances over 50 variables, and (c-d) are on benchmarks}\label{fig:expriment}
\end{figure*}

First, we compared the sizes of \ROBDDC{\widehat{i}}s ($0 \le i \le 20$) which represent random 3-CNF formulas over 50 variables, where each instance has 50, 100, 150 or 200 clauses. Figure \ref{fig:expriment}a depicts the experimental results. Each point is the mean value obtained over 100 instances with the same parameters. The experimental results show that the size of \ROBDDC{\widehat{i}} is normally not smaller than that of its equivalent \ROBDDC{\widehat{i+1}}, which is in accordance with the previous succinctness results.

Second, we implemented two conjoining algorithms of \ROBDD{} and \ROBDDL{\infty} \cite{PhDThesis} to conjoin two \ROBDDC{\widehat{0}}s and \ROBDDC{\widehat{1}}s, respectively. Note that a single conjunction is normally performed very fast. Taking into account the fact that bottom-up compilation of a CNF formula can be viewed as just performing conjunctions, we compare the bottom-up compiling time instead. Figure \ref{fig:expriment}b depicts the experimental results. Here we also used random instances with the same parameters as Figure \ref{fig:expriment}a, except that the numbers of clauses vary from 10 to 250. The experimental results show that for most instances, the time efficiency of conjoining \ROBDDC{\widehat{1}}s outperformed that of conjoining \ROBDDC{\widehat{0}}s, which accords with the rapidity results.

Last, we compared the sizes of \ROBDDC{\widehat{\infty}} with thoses of SDD (d-DNNF) on five groups of benchmarks: emptyroom, flat100 (flat200), grid, iscas89, and sortnet. We used a state-of-the-art compiler Dsharp \cite{Dsharp} to generate d-DNNF, and the SDD compiler in \cite{SDD-package} to generate SDD. Note that since the SDD compiler had some difficulty in compiling flat200, we used flat100 instead. Figure \ref{fig:expriment}c (\ref{fig:expriment}d) depicts the results both the \ROBDDC{\widehat{\infty}} and SDD (d-DNNF) compilers compiled in one hour. Due to space limit and readability consideration, we remove point $(9, 4)$ from Figure \ref{fig:expriment}c and then all other sizes are in the range of $[10^2, 10^6]$. Also, we remove points $(1192941, 82944)$, $(9, 3)$ and $(20104922, 6422)$ from Figure \ref{fig:expriment}d. Note that Figure \ref{fig:expriment}d has more points than Figure \ref{fig:expriment}c, since the compiling efficiency of SDD is normally lower than that of \ROBDDC{\widehat{\infty}} and d-DNNF. The experimental results show that the space-efficiency of \ROBDDC{\widehat{\infty}} is comparable with that of d-DNNF and SDD. Note that from the theoretical aspect, we can show that the succinctness relation between \ROBDDC{\widehat{i}} ($i > 0$) and SDD is incomparable, which accords with the experimental results.

\section{Related Work}
\label{sec:relat}

This study is closely related to three previous KC languages which also augment \BDD{} with decomposition.

First, \cite{Mateescu:etal:08} proposed a relaxation of \ROBDD{} called \emph{AND/OR Multi-value Decision Diagram} by adding tree-structured $\AND$-decomposition and ranking $\DEC$-vertices on the same tree-structured order. It is easy to see that for an AND/OR BDD (i.e., \AOBDD{}), if we remove all $\AND$-vertices with only one child, the result is an \OBDDC{}. And it is easy to show that \AOBDD{} is strictly less succinct than \ROBDDC{\widehat{\infty}}. In addition, \AOBDD{} is incomplete for non-chain trees.

Second, \cite{KAIS12} proposed a language called \emph{OBDD with implied literals} (\OBDDL{}) by associating each non-false vertex in OBDD with a set of implied literals, and then obtained a canonical subset called \ROBDDL{\infty} by imposing reducedness and requiring that every internal vertex has as many as possible implied literals. They designed an algorithm called L2Inf which can transform \OBDDL{} into \ROBDDL{\infty} in polytime in the size of input, and another algorithm called Inf2ROBDD which can transform \ROBDDL{\infty} into \ROBDD{} in polytime in the size of output. Obviously, each non-false vertex in \OBDDL{} can be seen as a $\AND_{1}$-vertex. Therefore, \OBDDL{} (\ROBDDL{\infty}) is equivalent to \OBDDC{1} (\ROBDDC{\widehat{1}}), and L2Inf and Inf2ROBDD are two special cases of \Decompose{} and \ConvertDown{}, respectively.

Last, \cite{Bertacco:Damiani:96} added the finest negatively-disjunctive-decomposition ($\NOR$-decomposition) into \ROBDD{} to propose a representation called \emph{Multi-Level Decomposition Diagram} (\MLDD{}). For completeness, $\NOT$-vertices are sometimes admitted. If we introduce both conjunctive and disjunctive decompositions into \ROBDD{}, then the resulting language will be equivalent to \MLDD{}. However, \cite{Bertacco:Damiani:96} paid little theoretical attention on the space-time efficiency of \MLDD{}. On the other hand, our empirical results show that there are little disjunctive decomposition in practical benchmarks.

\section{Conclusions}
\label{sec:concl}

The main contribution of this paper is a family of canonical representations, the theoretical evaluation of their properties based some previous criteria and a new criterion, and the experimental verification of some theoretical properties. Among all languages, \ROBDDC{\widehat{\infty}} has the best succinctness and rapidity. It seems to be the optimal option in the application where full compilation is adopted. However, it seems very time-consuming to directly compute the finest decomposition of a knowledge base since there normally exist too many possibilities of decomposition. Therefore, in the application where partial compilation is adopted (e.g., importance sampling for model counting \cite{Gogate:Dechter:11,Gogate:Dechter:12}), one may need other languages whose decompositions are relatively easy to be captured, for example, \ROBDDC{\widehat{1}} whose decompositions can be computed using SAT solver as an oracle \cite{KAIS12}. The second main contribution of the paper is the algorithms which perform logical operations or transform one language into another. These algorithms provide considerable potential to develop practical compilers for \ROBDDC{\widehat{i}}. Intrinsically, \ROBDDC{\widehat{i}} can be seen as a data structure which relax the linear orderedness of \ROBDD{} to some extent, and thus a future direction of generalizing this work is to exploit $\AND_{i}$-decomposition to relax the v-tree-structured order of SDD, which has the potential to identify new canonical languages with more succinctness than both \ROBDDC{\widehat{i}} and SDD.

\section*{Acknowledgments}
We would thank Arthur Choi for providing their 32-bit SDD package. This research is supported by the National Natural Science Foundation of China under grants 61402195, 61133011 and 61202308, and by China Postdoctoral Science Foundation under grant 2014M561292.

\bibliography{references,mypublications,softwares}
\bibliographystyle{named}

\end{document}